\documentclass[11pt,a4paper]{article}
\usepackage[hyperref]{acl2019}
\usepackage{times}
\usepackage{latexsym}
\usepackage{graphicx}

\usepackage{url}
\usepackage{placeins}
\usepackage{xcolor}

\aclfinalcopy 

\newcommand{\full}{Multi-Domain End-to-End Dialog System Platform}
\newcommand{\short}{ConvLab}

\title{\short: \full}

\author{Sungjin Lee$^\dag$ \quad Qi Zhu$^\ddag$ \quad Ryuichi Takanobu$^\ddag$ \quad Xiang Li$^\ddag$ \quad Yaoqin Zhang$^\ddag$ \quad Zheng Zhang$^\ddag$ \\ \textbf{Jinchao Li$^\dag$} \quad \textbf{Baolin Peng$^\dag$} \quad \textbf{Xiujun Li$^\dag$} \quad \textbf{Minlie Huang$^\ddag$} \quad \textbf{Jianfeng Gao$^\dag$} \\
  $^\dag$Microsoft Research, USA \quad $^\ddag$Tsinghua University, China \\
  $^\ddag${\small \tt \{zhu-q18,gxly15,zhangyq17,z-zhang15\}@mails.tsinghua.edu.cn} \\
  $^\dag${\small \tt \{sule,jincli,xiul,jfgao\}@microsoft.com} \quad 
  $^\ddag${\small \tt aihuang@tsinghua.edu.cn}
}

\date{}

\begin{document}
\maketitle
\begin{abstract}
We present \emph{\short}, an open-source multi-domain end-to-end dialog system platform, that enables researchers to quickly set up experiments with reusable components and compare a large set of different approaches, ranging from conventional pipeline systems to end-to-end neural models, in common environments. \short{} offers a set of fully annotated datasets and associated pre-trained reference models. As a showcase, we extend the MultiWOZ dataset with user dialog act annotations to train all component models and demonstrate how \short{} makes it easy and effortless to conduct complicated experiments in multi-domain end-to-end dialog settings. 
\end{abstract}

\section{Introduction}
Despite decades of research on dialog and increasingly large amounts of (annotated) dialog datasets, it is still challenging for any team who is new to the area to quickly develop a reasonable baseline system for task-oriented dialog due to the lack of a well-structured, easy-to-use open-source system that allows researchers to build and evaluate dialog bots. \short{} is aimed to fill the gap. \short{} is an open-source multi-domain end-to-end dialog system that allows researchers to automatically train dialog models, build and evaluate task-completion dialog bots. Such open-source systems have been instrumental in many AI research breakthroughs. For example, among many, Moses~\cite{koehn2007moses}, HTK~\cite{young2002htk} and CoreNLP~\cite{manning2014stanford} have been widely used to facilitate subsequent research in machine translation, speech recognition and natural language processing, respectively.

\short{} consists of a rich set of modeling tools and runtime engines for building task-oriented bots of different types, and an end-to-end evaluation platform. There are roughly two architectures of dialog systems~\cite{gao2019neural}: (1) modular architecture (the first layer in Figure~\ref{fig:agent}), consisting of natural language understanding (NLU), dialog state tracker (DST), dialog policy (POL) and natural language generation (NLG) components; and (2) fully end-to-end neural architecture (the last layer in Figure~\ref{fig:agent}) to minimize laborious hand-coding and error propagation down the pipeline. There also have emerged some models in-between~\cite{wen2016network,mrkvsic2016neural}. Due to the wide range of approaches and different metrics used in prior studies, it's been impracticable to perform a rigorous comparative study under the same condition. \short{} is the first dialog research platform that covers a full range of trainable statistical models with fully annotated datasets, differing from previous toolkits whose focus is largely concentrated on the system policy component while other components are mostly limited to pre-fixed baseline models~\cite{ultes2017pydial,miller2017parlai,li2018microsoft}.

There is also an increasing interest in building bots that seamlessly intertwine multiple sub-domains to accomplish high-level user goals 
~\cite{peng2017composite,budzianowski2018multiwoz}. The development of multi-domain dialog system adds additional complexities to both data collection and annotation, and the models for dialog system components. For the former, \citet{budzianowski2018multiwoz} collected the MultiWOZ dataset, a dialog corpus with dialogs ranging over multiple domains for the trip information setting,
whereas there is no open-platform yet that 
is designed to handle multi-domain, multi-intent phenomena.
To foster multi-domain dialog research, \short{} features the MultiWOZ task and offers a complete set of reference models ranging from individual components to end-to-end models that are trained on the MultiWOZ data with additional annotation for user dialog acts which is missing from the original MultiWOZ dataset. 
Furthermore, \short{} will be the standard platform for the multi-domain end-to-end task-completion dialog track in DSTC8\footnote{\url{https://sites.google.com/dstc.community/dstc8/home}}.

Finally, to support end-to-end evaluation, \short{} offers two complementary modules: Amazon Mechanical Turk integration for human evaluation and simulated users for automated evaluation. For user simulation, \short{} provides both rule-based simulators and data-driven simulators. As data-driven user simulation recently gains more traction, \short{} makes another contribution as a research platform for advancing user simulation technologies. 


The summary of the unique contributions of \short{} is:
\begin{itemize}
\itemsep-0.2em 
    \item To the best of our knowledge, \short{} is the first open-source multi-domain end-to-end dialog system that covers a full range of trainable statistical models with associated annotated datasets.
    \item \short{} provides a rich set of tools and recipes to develop dialog systems of different types, enabling researchers to compare widely different approaches under the same condition. 
    \item \short{} provides end-to-end evaluation via both human and simulators.
    \item We are organizing DSTC8 and releasing \short{} to public.
\end{itemize}

\begin{figure}[t!]
  \centering
  \includegraphics[width=0.95\linewidth]{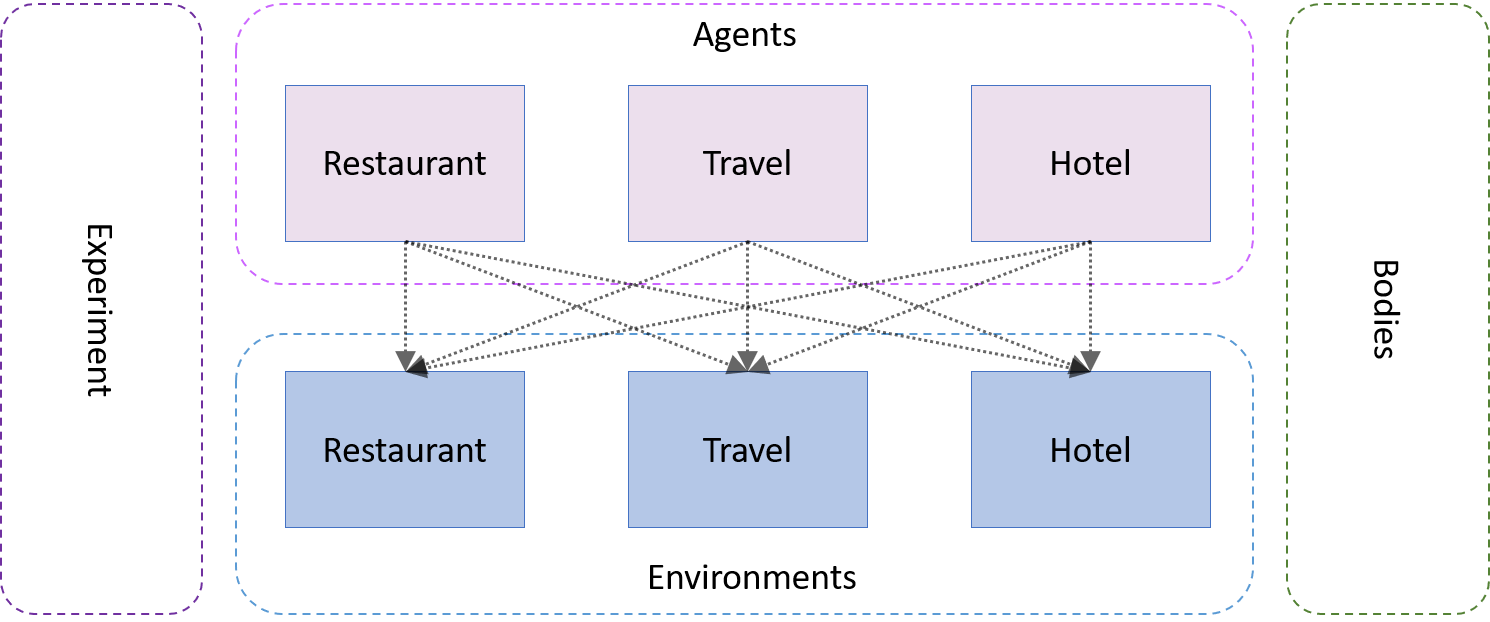}
  \caption{Overall design of ConvLab.}
  \label{fig:overall}
\end{figure}

\section{\short}
This section details the design of \short{} and its flexibility to support a wide range of experiments.  

\subsection{Overall Design}
At a high level, to support flexible architectures for multi-domain dialog, \short{} embraces the Agents-Environments-Bodies (AEB) design (illustrated in Figure~\ref{fig:overall}) with the following semantics~\cite{kenggraesser2017slmlab}: 
\begin{description}
\itemsep-0.2em 
\item[Agent] an instance of dialog agent.
\item[Environment] an instance of user simulator or human evaluation component.
\item[Body] an incarnation of an agent in the environment -- each body stores data that is specific to the associated agent and environment (indicated by the edges in Figure~\ref{fig:overall}): states, actions, rewards, done flags.
\end{description}

With the AEB design, besides the usual single agent and single environment setting, a variety of advanced research experiments, such as multi-agent learning, multi-task learning and role-play, can be conducted without requiring specialized code for each case.

\begin{description}
\itemsep-0.2em 
\item[Multi-agent learning] A centralized agent maps the joint observation of all domains to a joint action. A major drawback of this approach is its exponential growth in the observation and actions spaces with the number of domains. One can address this intractability by factoring the centralized spaces into multi-agent systems (including hierarchical reinforcement learning agents). For example, in Figure~\ref{fig:overall}, the centralized agent \emph{Travel} can be decomposed into two separate domain agents \emph{Restaurant} and \emph{Hotel}.
\item[Multi-task learning]  An agent can have multiple bodies in different environments with the goal of transfer learning. For example, any agent in Figure~\ref{fig:overall} can have its bodies not only in the corresponding environment but also in other environments to learn common knowledge across multiple domains. For example, in Figure~\ref{fig:overall}, each agent can learn from all available environments.
\item[Role play] Recently, there have been an increasing interest in leveraging self-play as an alternative way of training reinforcement learning agents~\cite{silver2017mastering}. Following the same spirit, for task-completion dialogs, one can devise a role play -- one agent plays the role of the system while the other agent as the user. Such a role play setting can be readily achieved by having two agents talk to each other though a round-robin environment.
\end{description}

For systematic comparison of agents and environments, and automated hyper-parameter search, \short{} makes use of SLM Lab~\cite{kenggraesser2017slmlab} and Ray\footnote{\url{https://github.com/ray-project/ray}} for the experiment component in Figure~\ref{fig:overall} which provides multi-level control layers, i.e.  Session, Trial and Experiment, and produces evaluation reports for each layer.
\begin{description}
\itemsep-0.2em 
\item[Session] Each session initializes the agents and environments and then runs for a pre-defined number of episodes.
\item[Trial] Each trial holds a fixed set of parameter values and runs multiple sessions with random seeds. The trial then analyzes the sessions and takes the average.
\item[Experiment] An experiment is a study where the hyper-parameters are treated as input variables, and the outcome is measured by task-specific metrics such as success rate and average reward. Search is then automatically conducted to find the hyper-parameters that yield best performance.
\end{description}

\short{} also helps avoid specifying complicated command line parameters and writing scripts by enabling users to control all relevant functionality via JSON configuration files. A configuration file specifies the model and its parameters for each component of the agent and environment for a given experiment. Thanks to the flexible configuration layer, researchers can build an array of different agents (Section~\ref{sec:agent}) and environments (Section~\ref{sec:env}) with only slight modifications in the configuration file. Some example configuration files are listed in Section~\ref{sec:demo}.


\subsection{Dialog Agent Configuration} \label{sec:agent}
\begin{figure}[h]
  \centering
  \includegraphics[width=0.5\linewidth]{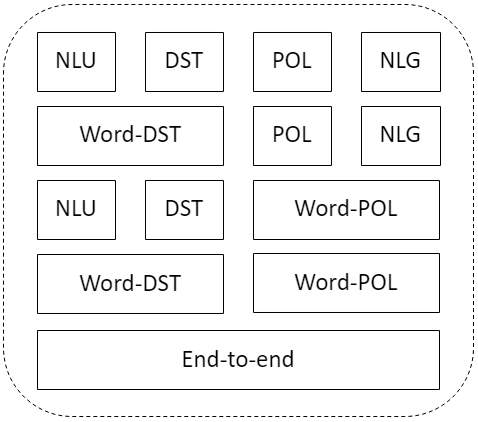}
  \caption{A dialog system configuration view.}
  \label{fig:agent}
\end{figure}


In Figure~\ref{fig:agent}, each layer represents a different way of constructing a dialog system. The first layer, for example, corresponds to the conventional pipeline architecture consisting of NLU, DST, POL and NLG. Recently, researchers have introduced some models that merge some of typical components such as word-level dialog state tracking, word-level dialog policy and end-to-end models, resulting in various possible combinations for building a dialog system as shown from the second layer in Figure~\ref{fig:agent}. However, comparison among these possibilities in an end-to-end setting has been largely overlooked, partly due to the burden of implementing all comparative systems. With \short, researchers can now focus on any particular component in Figure~\ref{fig:agent} while testing the algorithm in an end-to-end setting by simply creating a configuration file with a specification of other components.

\subsection{Environment Configuration}
\label{sec:env}
\begin{figure}[h]
  \centering
  \includegraphics[width=0.55\linewidth]{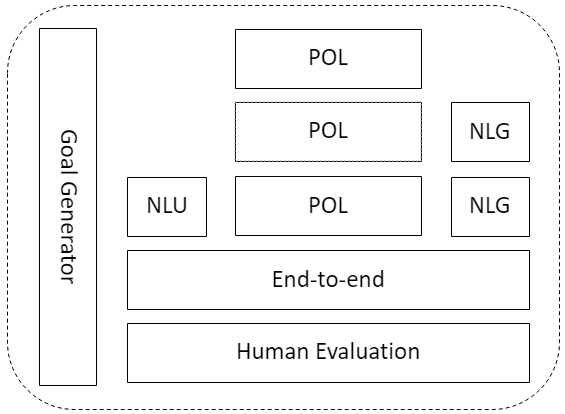}
  \caption{An environment configuration view.}
  \label{fig:env}
\end{figure}

As shown in Figure~\ref{fig:env}, there are also many different ways of combining some components to build an environment. For example, the first layer corresponds to a user simulator operating at the dialog act level which is the typical setting of prior works focusing on reinforcement learning algorithms for dialog policy optimization. As with dialog agent, there are recent attempts on end-to-end approaches to avoid requiring expensive annotation~\cite{kreyssig2018neural}. For human evaluation, \short{} also provides an integration of crowd source platform such as Amazon Mechanical Turk\footnote{\short{} makes use of ParlAI's MTurk library (\url{http://parl.ai/static/docs/mturk.html}).} as shown in the last layer. 

\subsection{Reference Models}
This section describes a set of reference models for each component that are available in the initial release. As we will keep adding new state-of-the-art models, the set of reference models available in \short{} will be extended. 

\paragraph{Natural Language Understanding} 
For natural language understanding, \short{} provides three reference models: Semantic Tuple Classifier (STC)~\cite{mairesse2009spoken}, OneNet~\cite{kim2017onenet} and Multi-intent LU (MILU). STC can handle multi-domain, multi-intent dialog acts but cannot detect out-of-vocabulary (OOV) values. While OneNet can capture OOVs, it cannot handle multi-intent dialog acts. Thus, \short{} offers a new MILU model which extends OneNet to process multi-intent dialog acts. For more details on MILU, please refer to the \short{} site.

\paragraph{Dialog State Tracking} 
The dialog state tracker is responsible for updating the belief state. \short{} provides a rule-based tracker similar to the baselines in DSTCs~\cite{williams2013dialog} that are adapted to handle multi-domain interactions.

\paragraph{Word-level Dialog State Tracking}
Word-level DSTs directly take system and user natural language as inputs and update dialog state. \short{} imports MDBT~\cite{ramadan2018large} model which jointly identifies the domain and tracks the belief states by utilizing the semantic similarity between dialog utterances and ontology terms. 

\paragraph{System Policy} 
For system policy, \short{} provides three classes of implementations: handcrafted policy, supervised learning policy and reinforcement learning policy. For reinforcement learning, \short{} supports a set of popular algorithms: DQN~\cite{mnih2013playing} and its variants, 
REINFORCE~\cite{williams1992simple}, PPO~\cite{schulman2017proximal} and its self-imitation variant
~\cite{oh2018self}
. For multi-domain dialog, \short{} initially offers centralized policies where the policy maps the joint observation of all domains to a joint action and will feature decentralized multi-agent approaches as well as hierarchical reinforcement learning approaches~\cite{peng2017composite}. 

\paragraph{Natural Language Generation} 
\short{} provides a template-based model and SC-LSTM~\cite{wen2015semantically} for natural language generation. Each model is able to take the multi-domain, multi-intent dialog acts as input.

\paragraph{Word-level Policy} 
Following~\citet{wen2016network}, word-level policy directly maps a context to response. \short{} imports the baseline implementation released for the benchmarking purpose by~\citet{budzianowski2018multiwoz}\footnote{\url{https://github.com/budzianowski/multiwoz}}. The baseline model extends a sequence-to-sequence model~\cite{sutskever2014sequence} with a dialog state encoding and a database query result encoding as additional features to the decoder.

\paragraph{User Policy}
For user policy, \short{} provides an agenda-based~\cite{schatzmann2007agenda} user model and data-driven approaches such as HUS
and its variational variants~\cite{gur2018user}. Similar to the system side, each model works at the dialog act level, and can be pipelined with NLU and NLG modules to construct a whole user simulator.

\paragraph{End-to-end Model} 
\short{} makes available two end-to-end dialog system models: Mem2Seq~\cite{madotto2018mem2seq} and Sequicity~\cite{lei2018sequicity}. To support multi-domain intents, Sequicity resets the belief span when the model predicts a topic shift between domains.

\section{Domains}
The initial release of \short{} offers two domains of differing complexity: MultiWOZ and Movie. 

\paragraph{MultiWOZ}
The main task of the MultiWOZ domain is to help a tourist in a various situations involving multiple sub-domains such as requesting basic information about attractions and booking a hotel room. Specifically, there are 7 sub-domains - \texttt{Attraction, Hospital, Police, Hotel, Restaurant, Taxi, Train}. The annotated data consists of 10,438 dialogs. The average number of turns are 8.93 and 15.39 for single and multi-domain dialogs, respectively. \short{} features additional annotations for user dialog acts and pre-trained reference models for all dialog system components and user simulators. Furthermore, \short{} provides a set of end-to-end neural dialog models that are trained on the data.

\paragraph{Movie}
\short{} imports the \texttt{Movie} domain from Microsoft Dialog Challenge~\cite{li2018microsoft}, encouraging researchers to continue working on the movie ticket booking task with enhanced tools. 
The annotated dataset consists of 2,890 dialogs, with approximately 7.5 turns per dialog on average. \short{} offers a complete reference set of models trained on the data for both agent and user simulator. 

We plan to add more domains such as the \texttt{Taxi} and \texttt{Restaurant} domains from Microsoft Dialog Challenge.

\section{Demo}
\label{sec:demo}
To demonstrate a glimpse of some working systems, this section presents two experiments: 1) comparison between NLU with rule-based DST and word-level DST; 2) comparison between rule-based policy with NLG and word-level policy.

\paragraph{Experiment 1}
Word-level DSTs often have shown higher performance than typical DSTs that take input from NLU~\cite{ramadan2018large,mrkvsic2016neural}, but none of prior works confirmed the performance improvement in an end-to-end setting. Thanks to the flexible configuration interface and pre-trained reference models, with \short, one can easily set up end-to-end experiments by simply modifying a few lines in the config files as listed in Table~\ref{tab:config}. While the overall accuracies of the rule-based DST and the word-level DST are 90.2\% and 89.7\%, respectively, the end-to-end task success rates are 69.05\% and 16.67\%. This clearly shows the gap between component-level performance and end-to-end performance. A detailed analysis on this is left for future work.

\paragraph{Experiment 2}
Though word-level policy obtains an increasing traction, most studies only report corpus-based metrics such as BLEU and pseudo-success rate (i.e. success means all requested attributes are answered). This makes it hard to compare such approaches with conventional policies that are typically evaluated with task success metrics. 
Due to the space limitations, we omit the experimental config which is largely the same as the config listed on the left column in Table~\ref{tab:config} except that the \texttt{policy} and \texttt{nlg} sections under the \texttt{agent} section are now replaced with a corresponding \texttt{word\_policy} section. While the reported pseudo-success rate on test data is 60.96\%, the success rate with a user simulator is 16.16\%. This is also much lower than 69.05\%, the performance of its counterpart with rule-based policy and NLG. Thus, there is huge room for improvement of the word-level policy in end-to-end settings.


\begin{table}[t!]
\tiny
    \centering
\begin{tabular}{l|l}
\hline
NLU and rule-based DST & Word-level DST \\ \hline
\begin{minipage}{1.5in}
\begin{verbatim}
{"multiwoz": {
  "agent": [{
   "name": "DialogAgent",
    "nlu": {
     "name": "OneNet"
    },
   "dst": {
    "name": "RuleDST"
   },
   "policy": {
    "name": "ExternalPolicy",
    "algorithm": {
     "name": "RulePolicy"
    }
   },
   "nlg": {
    "name": "TemplateNLG",
    "is_user": false
   }      
  }],
  "env": [{
   "name": "multiwoz",
   "nlu": {
    "name": "OneNet"
   },
   "policy": {
    "name": "UserPolicyAgenda"
   },
   "nlg": {
    "name": "TemplateNLG",
    "is_user": true
   }      
   "max_t": 40,
   "max_tick": 20000,
  }],
  "body": {
   "product": "outer",
   "num": 1
}}} 
\end{verbatim}
\end{minipage} &  
\begin{minipage}{1.3in}
\begin{verbatim}
{"multiwoz": {
  "agent": [{
   "name": "DialogAgent",
   "word-dst": {
    "name": "MDBT"
   },
   "policy": {
    "name": "ExternalPolicy",
    "algorithm": {
     "name": "RulePolicy"
    }
   },
   "nlg": {
    "name": "TemplateNLG",
    "is_user": false
   }      
  }],
  "env": [{
   "name": "multiwoz",
   "nlu": {
    "name": "OneNet"
   },
   "policy": {
    "name": "UserPolicyAgenda"
   },
   "nlg": {
    "name": "TemplateNLG",
    "is_user": true
   }      
   "max_t": 40,
   "max_tick": 20000,
  }],
  "body": {
   "product": "outer",
   "num": 1
}}}
\end{verbatim}
\end{minipage} \\  \hline
\end{tabular}
    \caption{Example configs for comparing a system using word-level DST (right) with one using NLU and rule-based DST (left).}
    \label{tab:config}
\end{table}

\section{Code and Resources}
The \short{} platform is publicly available from \url{http://convlab.github.io}.\footnote{The site will become accessible after a legal process is done.} Datasets and other resources such as tutorials and documentations can be found from the site.  

\section{Conclusion}
We presented \emph{ConvLab}, an open-source multi-domain end-to-end dialog system platform, that enables researchers to quickly set up experiments and compare different approaches without much effort. We will keep extending \short{} by adding new state-of-the-art models going forward. The multi-domain end-to-end task completion dialog track in DSTC8 will employ \short{} as the challenge platform, giving rise to a reference use case. 

\bibliography{acl2019}
\bibliographystyle{acl_natbib}


\end{document}